\documentclass[conference]{IEEEtran}

\usepackage{amsmath}
\usepackage{paralist}
\usepackage{tikz}
\usetikzlibrary{shapes.geometric, arrows}
\usepackage{cite}
\usepackage{hyperref}
\usepackage{subcaption}
\usepackage{graphicx}
\usepackage[most]{tcolorbox}
\usepackage{fontawesome}

\ifCLASSINFOpdf
\else
\fi

\begin{document}

\title{Neurosymbolic AI for Enhancing Instructability \\ in Generative AI}


% \author{
% \IEEEauthorblockN{Amit Sheth}
% \IEEEauthorblockA{University of South Carolina\\
% Columbia, SC, 29208, USA}
% \and
% \IEEEauthorblockN{Vishal Pallagani}
% \IEEEauthorblockA{University of South Carolina\\
% Columbia, SC, 29208, USA}
% \and
% \IEEEauthorblockN{Kaushik Roy}
% \IEEEauthorblockA{University of South Carolina\\
% Columbia, SC, 29208, USA}
% }

\author{
    Amit Sheth, Vishal Pallagani, Kaushik Roy\\
    Artificial Intelligence Institute of South Carolina\\
    University of South Carolina, Columbia, SC, 29208, USA
}

\maketitle

\begin{abstract}
Generative AI, especially via Large Language Models (LLMs), has transformed content creation across text, images, and music, showcasing capabilities in following instructions through prompting, largely facilitated by instruction tuning. Instruction tuning is a supervised fine-tuning method where LLMs are trained on datasets formatted with specific tasks and corresponding instructions. This method systematically enhances the model's ability to comprehend and execute the provided directives. Despite these advancements, LLMs still face challenges in consistently interpreting complex, multi-step instructions and generalizing them to novel tasks, which are essential for broader applicability in real-world scenarios. This article explores why neurosymbolic AI offers a better path to enhance the instructability of LLMs. We explore the use a symbolic task planner to decompose high-level instructions into structured tasks, a neural semantic parser to ground these tasks into executable actions, and a neuro-symbolic executor to implement these actions while dynamically maintaining an explicit representation of state. We also seek to show that neurosymbolic approach enhances the reliability and context-awareness of task execution, enabling LLMs to dynamically interpret and respond to a wider range of instructional contexts with greater precision and flexibility.
\end{abstract}

\IEEEpeerreviewmaketitle

\section{Introduction}

\begin{figure}[htbp]
    \centering
    \includegraphics[width=\linewidth]{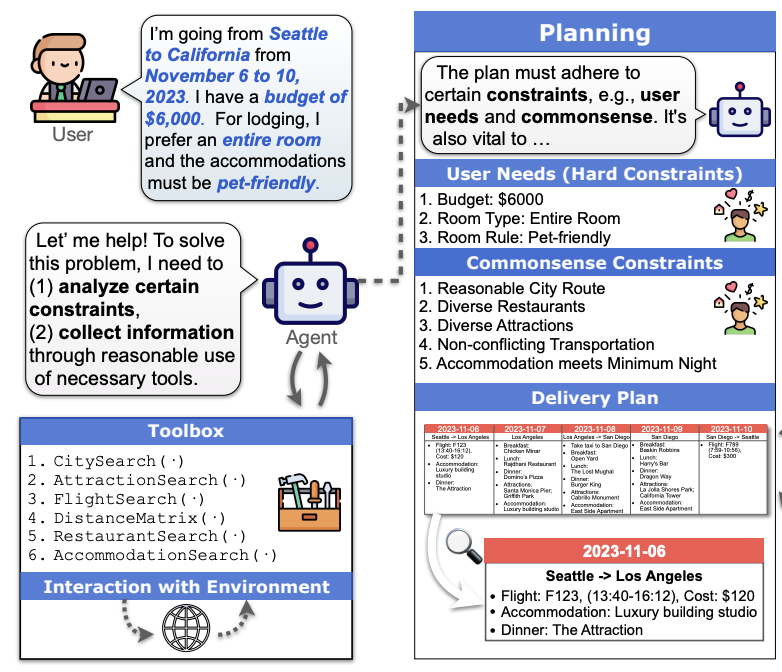}
    \caption{A sample instruction from the TravelPlanner dataset where a complex multi-step instruction from a user is illustrated, which requires decomposing the instruction into executable actions. State-of-the-art LLMs could not handle such complex instructions, with GPT-4 successfully producing a plan that meets all the constraints for only 0.6\% instructions, while all other LLMs fail to complete any tasks \cite{xie2024travelplanner}.}
    \label{fig:travel-planner}
\end{figure}

Large language models such as OpenAI’s GPT-4 that is used by ChatGPT and Meta’s Llama have demonstrated unprecedented capabilities in following natural language instructions through prompting. These models can be adapted to a broad spectrum of tasks by simply receiving appropriately structured prompts, showcasing their versatility and utility across various domains. For example, LLMs can successfully generate programming code from descriptions of desired functionalities or craft well-structured essays based on outlined themes, showcasing their adaptability and precision in tasks with clear, well-defined goals. However, despite their impressive performance, the current teaching paradigm in LLMs encounters several challenges. These include 
\begin{inparaenum}[(i)]
    \item the handling of complex, multi-step instructions, 
    \item inconsistencies in interpreting instructions due to ambiguous language or contextual nuances, 
    \item a limited ability to generalize to novel task compositions that deviate from trained examples, and 
    \item a lack of explicit reasoning mechanisms that can delineate and manage the execution processes involved in following instructions.
\end{inparaenum}
A major drawback is evident in planning travel itineraries where LLMs often fail to consider real-world constraints and user preferences effectively. For instance, when tasked with creating a multi-city travel plan as shown in Figure \ref{fig:travel-planner}, LLMs might generate a sequence that is theoretically correct but impractical, such as suggesting flights that don't exist or ignoring necessary travel recovery times and local conditions \cite{xie2024travelplanner}.

\begin{figure*}[htbp]
    \centering
    \includegraphics[width=\linewidth]{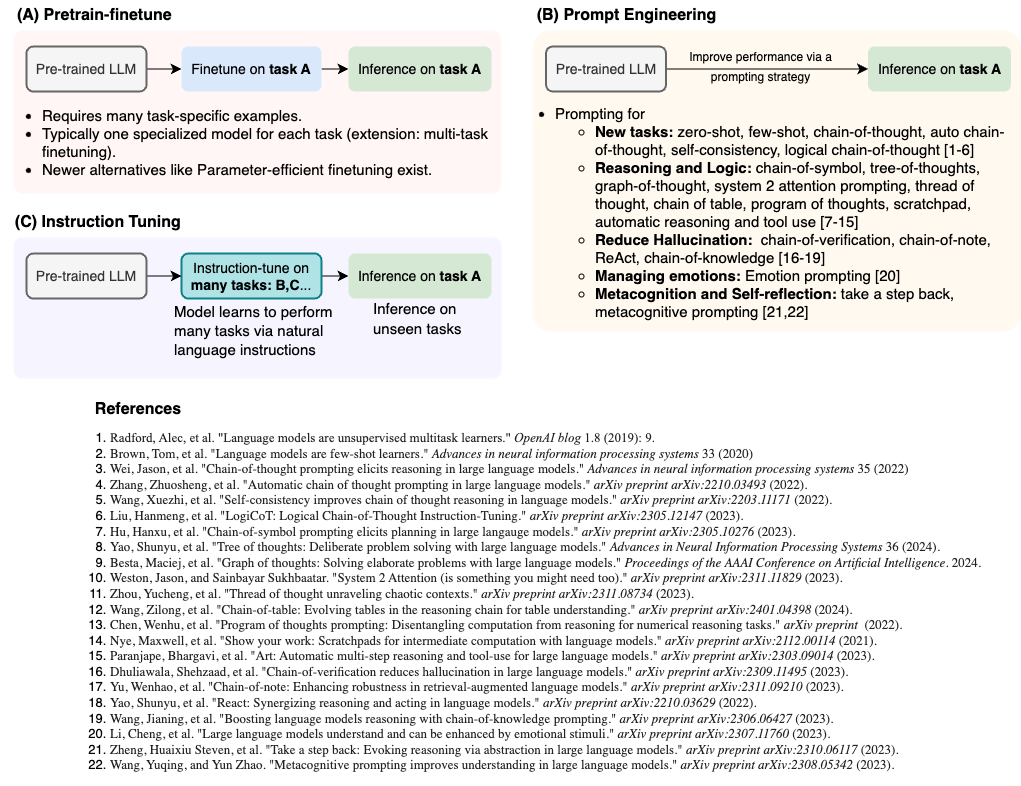}
    \caption{Notable difference between finetuning, prompting, and instruction tuning for LLMs.}
    \label{fig:llm-adaptation}
\end{figure*}

A significant part of these challenges stems from the nature of instruction tuning, the prevalent method for enhancing LLMs' ability to follow instructions. Figure \ref{fig:llm-adaptation} shows the difference between finetuning, prompting, and instruction tuning as used today for better understanding. Instruction tuning involves supervised fine-tuning of models on datasets specifically designed with pairs of tasks and corresponding instructions. While this approach effectively improves task-specific performance, it relies heavily on the availability of large volumes of representative demonstrations. This dependency can limit the model’s performance on rare or novel tasks due to the low probability of such events occurring in the training data. Moreover, instruction-tuned models often struggle to extrapolate learned behaviors to new contexts, indicating a gap in their ability to generalize beyond their immediate training environments \cite{kung2023models}. If the learned model does not have concepts relevant to the instruction, the instruction is ineffective or can create noise. Additionally, the pre-training objectives of these models, such as next token prediction or fill-in-the-middle, do not adequately ensure real-world grounding, leading to outputs that are textually coherent but practically infeasible in complex real-world applications.

To address these limitations, we propose adapting techniques from neurosymbolic AI \cite{sheth2023neurosymbolic}. We seek to enhance the instructability of LLMs by integrating the strengths of both neural networks and symbolic reasoning. Neurosymbolic AI seeks to combine the flexibility and pattern recognition capabilities of neural networks with the systematic, rule-based reasoning and generalization properties of symbolic AI. This hybrid approach promises to mitigate the weaknesses inherent in purely neural systems, particularly in handling complex tasks that require structured reasoning and dynamic adaptability.

In this article, we outline a neurosymbolic framework designed to achieve robust and generalizable instructability in generative AI. Our framework consist ofthree key components:
\begin{enumerate}
    \item \textbf{Symbolic Task Planner}: This module breaks down complex instructions into structured, manageable tasks, allowing for a clearer and more systematic approach to task execution.
    \item \textbf{Neural Semantic Parser}: This component grounds the decomposed tasks into specific, executable actions, translating abstract instructions into concrete operations that can be dynamically adapted to the task at hand.
    \item \textbf{Neurosymbolic Executor}: Operating with an explicit representation of the current state, this executor implements the actions while continuously updating its state awareness, enabling real-time adjustments and decision-making based on the evolving context.
\end{enumerate}

% \begin{figure*}[htbp]
%     \centering
%     \begin{subfigure}[b]{0.39\textwidth}
%         \centering
%         \includegraphics[width=0.38\textwidth]{figures/robotic-gripper.png}
%         \caption{A hierarchical task sequence for the \texttt{fetch object} process executed by a robotic gripper, illustrating how the steps are structured and represented within a process knowledge graph.}
%         \label{fig:first_image}
%     \end{subfigure}

%     \begin{subfigure}[b]{0.59\textwidth}
%         \centering
%         \includegraphics[width=0.58\textwidth]{figures/medical-pk.png}
%         \caption{An agent leveraging process knowledge for generating safe and medically accurate follow-up questions \cite{9889132}.}
%         \label{fig:second_image}
%     \end{subfigure}
    
%     \caption{Illustrations of hierarchical task ordering, capturing subtasks and their integration into higher-level tasks within a process knowledge graph. Figure \ref{fig:first_image} demonstrates a robotic gripper's \texttt{fetch object} task, while Figure \ref{fig:second_image} shows the use of process knowledge to enhance safety in conversational agents.}

%     \label{fig:main_figure}
% \end{figure*}

\begin{figure*}[htbp]
    \centering
    \begin{subfigure}[b]{0.4\textwidth}
        \centering
        \includegraphics[width=0.98\textwidth]{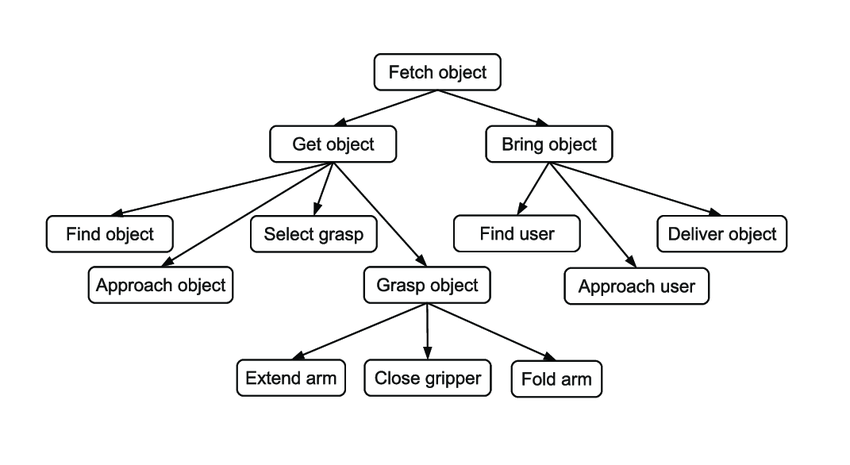}
        \caption{A hierarchical task sequence for the \texttt{fetch object} process executed by a robotic gripper, illustrating how the steps are structured and represented within a process knowledge graph.}
        \label{fig:first_image}
    \end{subfigure}
    \hfill
    \begin{subfigure}[b]{0.58\textwidth}
        \centering
        \includegraphics[width=0.98\textwidth]{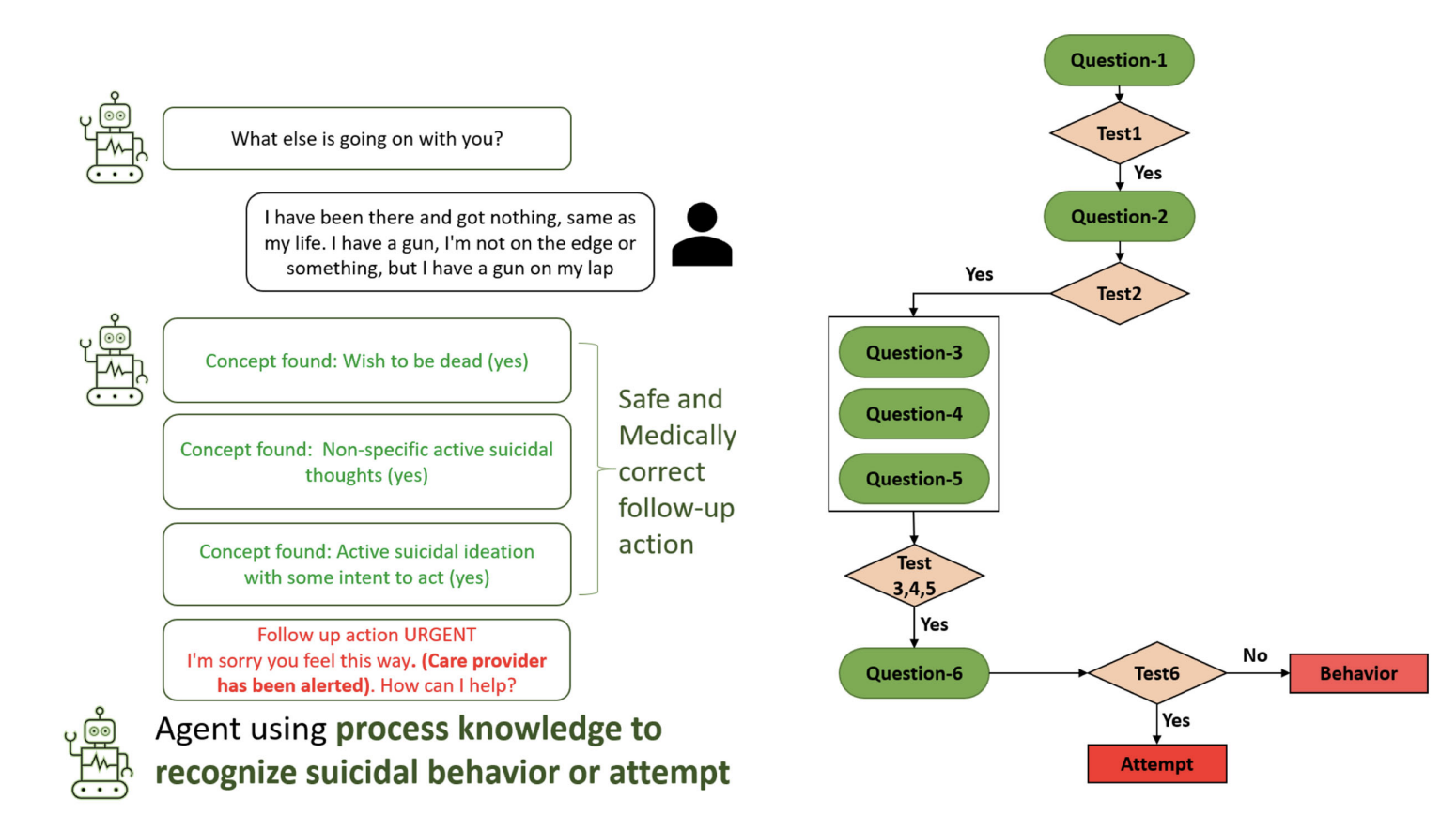}
        \caption{An agent leveraging process knowledge for generating safe and medically accurate follow-up questions \cite{9889132}.}
        \label{fig:second_image}
    \end{subfigure}
    
    \caption{Illustrations of hierarchical task ordering, capturing subtasks and their integration into higher-level tasks within a process knowledge graph. Figure \ref{fig:first_image} demonstrates a robotic gripper's \texttt{fetch object} task, while Figure \ref{fig:second_image} shows the use of process knowledge to enhance safety in conversational agents.}
    \label{fig:main_figure}
\end{figure*}

This framework delivers improved accuracy and consistency in following detailed multi-step instructions but will also show enhanced capability to generalize to new and diverse scenarios. This approach, therefore, sets the stage for a more reliable and versatile application of generative AI models in real-world tasks, where unpredictability and complexity are the norms rather than exceptions.

\section{Instructability So Far}

Traditionally, instructability in intelligent systems was rooted in symbolic AI, which utilized explicit, rule-based programming and logical frameworks to provide deterministic guarantees for instruction execution\cite{macdonald1991instructable}. As the LLMs have evolved, use and support for instructability has evolved. Currently, LLMs are widely used as chatbots, popularized by OpenAI’s ChatGPT. This trend can be traced back to the world’s first chatbot, Eliza, developed in the 1960s. With the increasing adoption of ChatGPT and similar LLM-based chatbots, the need for instructability has become more pronounced as people rely on these chatbots to obtain information across a wide range of domains. 

One of the major challenges with LLMs is aligning with users' objectives, primarily due to a mismatch between model's training goals and the users' needs: LLMs are typically trained on minimizing the contextual word prediction error on large corpora, while users want the model to “follow their instructions helpfully and safely” \cite{radford2019language, brown2020language}. To address this mismatch, instruction tuning was proposed, which involves further training LLMs using (INSTRUCTION, OUTPUT) pairs, where INSTRUCTION denotes the human instruction for the model, and OUTPUT denotes the desired output that follows the INSTRUCTION. The main challenges of instruction tuning are: (a) crafting high-quality instructions that cover target behaviors is challenging due to limited quantity, diversity, and creativity in existing datasets, (b) instruction tuning improves on tasks that are heavily represented in the training dataset, raising concerns about its effectiveness on less represented tasks \cite{gudibande2023false}, and (c) instruction tuning captures only surface-level patterns and styles rather than comprehending and learning the task; limiting required understanding to perform a specific task \cite{kung2023models} as seen in Figure \ref{fig:travel-planner}. 

Instruction tuning is essentially a rebranded version of well-known supervised fine-tuning, and as such, it shares many common issues with training a model on large datasets and expecting reliable and consistent performance on tasks that require more advanced, system 2 thinking abilities \cite{sheth2023neurosymbolic}. However, instruction tuning remains a very active area of research, and we encourage readers to refer to this \href{https://github.com/xiaoya-li/Instruction-Tuning-Survey}{Github} repository for the latest developments.

Following instructions is critically important in applications or domains such as planning-like tasks \cite{srivastava2024case}, robotics, and mental health. Observations across these applications indicate that LLMs frequently fail to accurately follow instructions to reach the user-desired goal state \cite{9889132, pallagani2023understanding, wu2024safety}. The outputs often exhibit inaccuracies, including hallucinations and significant issues with grounding affordances, reflecting a disconnect between the models' outputs and real-world applications \cite{pallagani2024prospects}. This disconnect arises because instructions often involve complex, multi-step processes that necessitate further decomposition and contextual understanding for LLMs to interpret and execute them accurately.

Domain-specific neurosymbolic approaches have been proposed to enhance plan generation \cite{fabiano2023plan,fabiano2023fast} and adherence to mental health guidelines \cite{9889132}. However, there remains a lack of a unified neurosymbolic framework aimed at improving the general-purpose instructability of LLMs. Developing such an architecture could significantly enhance LLMs' consistency, reliability, explainability, and safety across various applications \cite{gaur2024building}.

\section{Towards Neurosymbolic Approach to Instructability}

In this section, we describe our proposed neurosymbolic instruction following framework.

\subsection{Symbolic Task Planner}
We propose employing a symbolic task planner that leverages Hierarchical Task Networks (HTNs) to decompose complex instructions into clearly defined, manageable subtasks. This planner is equipped with an extensive library of task schemas, which serve as blueprints for decomposing specific types of tasks. For this, we can use process knowledge graphs which capture detailed task-specific guidelines or protocols relevant to solving the problems, as shown in Figure \ref{fig:main_figure}. Such guidelines exist for many applications, for example, in case of assisting a clinician, the process may be prescribed in the clinical practice guidelines (see Figure \ref{fig:second_image}. By recursively breaking down high-level instructions (see Figure \ref{fig:task_planner}), the planner translates them into finer, executable primitive actions, guided by predefined planning rules and constraints inherent to the HTNs. The integration of process knowledge graphs is crucial as they provide the necessary contextual and operational knowledge for each task. These graphs ensure that each decomposed action is not only logically consistent but also enriched with relevant domain-specific information.

\begin{figure}[h]
\centering
\begin{tcolorbox}[colback=blue!5!white, colframe=blue!75!black, title={Symbolic Task Planner Process}]
The task planner receives the high-level instruction to \textit{\textcolor{blue}{``Plan a 10-day cultural tour of Japan, including visits to Tokyo, Kyoto, and Nara during cherry blossom season.''}} It decomposes this into manageable subtasks:
\begin{itemize}
    \item[\faCalendarCheckO] \textbf{\textcolor{blue}{Determine optimal travel dates}} within cherry blossom season.
    \item[\faMapMarker] \textbf{\textcolor{blue}{Identify key cultural sites}} and events in Tokyo, Kyoto, and Nara.
    \item[\faListAlt] \textbf{\textcolor{blue}{Outline a day-by-day itinerary}} balancing travel time and site visits.
\end{itemize}
This decomposition helps in creating a structured framework for further detailing and execution.
\end{tcolorbox}
\caption{Decomposition of a high-level instruction by the Symbolic Task Planner}
\label{fig:task_planner}
\end{figure}

\subsection{Neural Semantic Parser}
Following the structured decomposition achieved by the planner, we employ a neural semantic parser, specifically fine-tuning a pre-trained compact language model for this purpose. This parser is tasked with translating the hierarchically organized subtasks and natural language instructions into a grounded representation of actions (see Figure \ref{fig:semantic_parser}) along with their requisite parameters and arguments. This translation process is crucial for converting the symbolic planner's output, which organizes tasks at a high conceptual level, into detailed, executable commands. By grounding these decomposed tasks into actionable language forms, the neural semantic parser acts as a critical bridge, transforming high-level linguistic constructs into precise, actionable outputs that are ready for execution. This step not only ensures that the instructions are executable but also maintains semantic fidelity to the original user intent, thereby enhancing the system’s ability to accurately follow complex instructions.

\begin{figure}[h]
\centering
\begin{tcolorbox}[colback=orange!5!white, colframe=orange!75!black, title={Neural Semantic Parser Process}]
The Neural Semantic Parser receives the subtask to \textit{\textcolor{orange}{``Identify key cultural sites and events in Tokyo, Kyoto, and Nara.''}} Here are the specific, executable actions it generates:
\begin{itemize}
    \item[\faMapMarker] \textbf{\textcolor{orange}{Locations Identified:}}
    \begin{itemize}
        \item Tokyo: \textbf{Action} - ``Book entry for Tokyo Skytree on April 3rd, 10:00 AM.''
        \item Kyoto: \textbf{Action} - ``Schedule visit to Kinkaku-ji on April 5th, 1:00 PM.''
        \item Nara: \textbf{Action} - ``Reserve participation in Nara Deer Park feeding event on April 6th, 3:00 PM.''
    \end{itemize}
    \item[\faCalendar] \textbf{\textcolor{orange}{Events Scheduled:}}
    \begin{itemize}
        \item ``Arrange for attendance at the Sakura Cherry Blossom Festival in Ueno Park, Tokyo on April 2nd.''
    \end{itemize}
    \item[\faTicket] \textbf{\textcolor{orange}{Ticketing Details:}}
    \begin{itemize}
        \item ``Secure online tickets for the Gion Matsuri parade in Kyoto, ensuring access on April 4th.''
    \end{itemize}
\end{itemize}
This transformation ensures each task is grounded into actionable and specific bookings or event participations, ready for execution by the neurosymbolic executor.
\end{tcolorbox}
\caption{Processing of one of the decomposed task obtained from Symbolic Task Planner by the Neural Semantic Parser}
\label{fig:semantic_parser}
\end{figure}

\begin{figure}[h]
\centering
\begin{tcolorbox}[colback=purple!5!white, colframe=purple!75!black, title={{Neurosymbolic Executor Process}}]
The Neurosymbolic Executor receives the grounded task \textit{\textcolor{purple}{``Book entry for Tokyo Skytree on April 3rd, 10:00 AM.''}} and processes:

\begin{itemize}
    \item[\faTicket] \textbf{\textcolor{purple}{Action Execution:}}
    \begin{itemize}
        \item Execute the booking action through the online ticketing system.
        \item Confirm and retrieve booking confirmation details.
    \end{itemize}
    \item[\faRefresh] \textbf{\textcolor{purple}{State Update and Monitoring:}}
    \begin{itemize}
        \item Update the travel itinerary state to include the confirmed booking.
        \item Monitor real-time updates for any changes to the booking status or Skytree event schedules.
    \end{itemize}
    \item[\faExchange] \textbf{\textcolor{purple}{Dynamic Adjustments:}}
    \begin{itemize}
        \item Adjust travel and visit plans if there are any disruptions or changes in the event schedule.
        \item Respond to weather conditions or other external factors affecting the visit day.
    \end{itemize}
\end{itemize}

\textbf{Output:} Successfully executed and dynamically managed booking for Tokyo Skytree, fully integrated into the overall itinerary with real-time adjustments prepared.
\end{tcolorbox}
\caption{Execution of a grounded task by the Neurosymbolic Executor}
\label{fig:neurosymbolic_executor}
\end{figure}

\subsection{Neurosymbolic Executor}
The neurosymbolic executor is the operational core of our framework, responsible for implementing the grounded instructions provided by the neural semantic parser. It maintains an explicit symbolic representation of the state, crucial for assessing and managing the ongoing changes within the execution environment. This executor integrates neural components, which are adept at perception and dynamic action execution, with robust symbolic reasoning mechanisms. Such integration enables precise tracking of state changes and effective management of control flow.

\begin{figure}
    \centering
    \includegraphics[width=\linewidth]{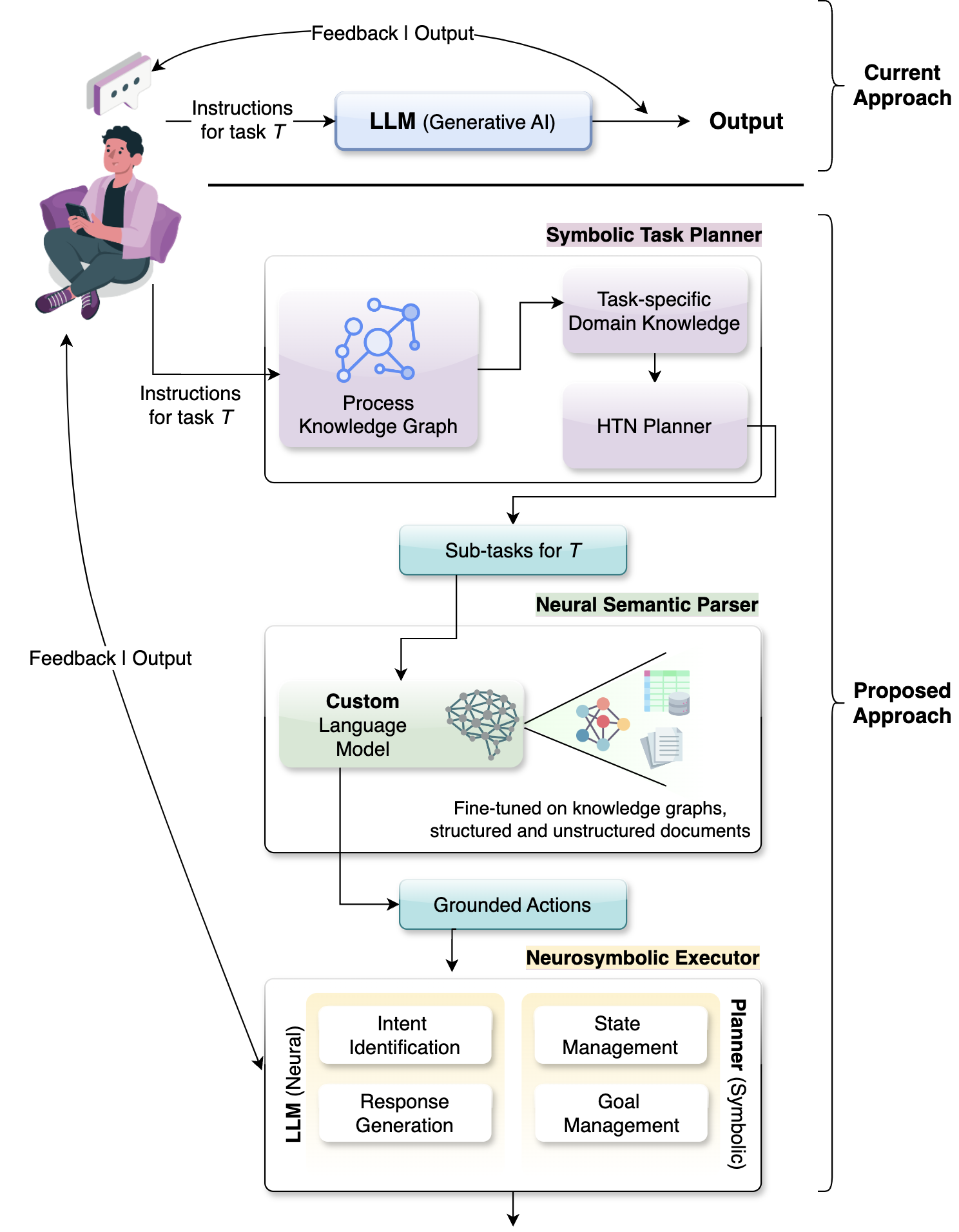}    
    \caption{Comparison between current approach to instruct LLMs to provide a response for a task and the proposed neurosymbolic approach for instructability.}
    \label{fig:nesy-instruct}
\end{figure}

The hybrid nature of this executor allows for dynamic adjustments, adapting in real-time to the complexities and unpredictability encountered during instruction execution. By leveraging both the predictive strengths of neural models and the deterministic nature of symbolic logic, the executor ensures a seamless and coherent execution process. This approach not only enhances the reliability and accuracy of following complex instructions but also supports the system's ability to handle interruptions, unexpected conditions, and varying contextual cues with high resilience and adaptability (see Figure \ref{fig:neurosymbolic_executor}). Figure \ref{fig:nesy-instruct} shows the comparison between current approach to instruct LLMs to perform a task and the proposed neurosymbolic instructability approach.

\section{Dependency of Instructability on Grounding and Alignment}

The LLMs fundamentally hinge on two critical capabilities: grounding and alignment. Grounding refers to the ability of LLMs to connect the language constructs used in their outputs with real-world entities and contexts. Alignment, on the other hand, involves the LLMs' ability to produce outputs that are not only contextually appropriate but also closely aligned with the users' intentions and ethical guidelines. Both grounding and alignment are pivotal for ensuring that LLMs can follow instructions in a way that meets the practical and ethical expectations of their human users.

\subsection{Grounding in LLMs}

Grounding in LLMs \cite{bajaj2024grounding} involves the translation of abstract linguistic representations into concrete, actionable entities and scenarios. This process is essential for LLMs as it impacts their ability to interpret and execute complex instructions accurately within a specified context. For example, in task-oriented applications such as navigating an environment or executing a series of physical actions, the LLM must understand and map its instructions onto the physical world (see Figure \ref{fig:grounding_alignment}). This requirement extends beyond simple recognition of terms; it necessitates an understanding of the relationships and interactions between various entities. Without effective grounding, LLMs are prone to generating outputs that, while linguistically correct, are made up, infeasible or irrelevant. The integration of knowledge graphs are kept up-to-date can significantly enhance grounding by providing a rich, interconnected database of real-world entities and their attributes, enabling LLMs to draw on a vast reservoir of structured information to better interpret and relate instructions to tangible, real-world applications.

\begin{figure}[htbp]
\centering
\begin{tcolorbox}[colback=white!10!white, colframe=blue!75!black, title=Instruction]
Plan a 10-day cultural tour of Japan, including visits to Tokyo, Kyoto, and Nara during cherry blossom season.
\end{tcolorbox}

\begin{tcolorbox}[colback=white!10!white, colframe=green!55!black, title=Grounding in Travel Planning]
Grounding involves linking abstract language instructions to tangible, specific real-world contexts and actions. Here’s how grounding is reflected in the travel planning task:
\begin{itemize}
    \item \textbf{Temporal Context:} Identifies the "cherry blossom season" as typically occurring from late March to early April, which directly affects the timing of the travel recommendations.
    \item \textbf{Cultural Significance:} Suggests specific cherry blossom viewing spots, like Ueno Park in Tokyo and the Philosopher's Path in Kyoto, grounding the travel experience in culturally significant activities.
    \item \textbf{Historical Sites:} Includes visits to historically significant sites such as Kyoto’s Kinkaku-ji and Nara’s Todai-ji, grounding the itinerary in Japan's rich historical context.
    \item \textbf{Travel Logistics:} Recommends booking Shinkansen (bullet train) tickets for efficient inter-city travel, grounding the plan in practical travel logistics within Japan.
\end{itemize}
\end{tcolorbox}
\caption{Illustration of Grounding in LLM-based Travel Planning}
\label{fig:grounding_alignment}
\end{figure}

\subsection{Alignment in LLMs}

Alignment involves ensuring that LLMs' actions and responses are effective, ethically sound, and aligned with user expectations (see Figure \ref{fig:alignment_travel_planning}). This aspect of LLM behavior is crucial for maintaining trust and reliability, particularly in sensitive applications such as healthcare, legal advice, or educational settings. Alignment ensures that the LLM's responses adhere to ethical standards and user-specific requirements, preventing scenarios where the model's behavior diverges from human values or produces harm \footnote{https://arxiv.org/pdf/2312.09928}. Effective alignment strategies involve technical measures, such as adjusting model parameters and training data, and policy measures, such as incorporating feedback loops that allow users to report and rectify misaligned behavior. A knowledge graph can further support alignment by directly embedding a structured understanding of ethical norms and user preferences into the model’s reasoning processes, providing a foundational layer that helps guide the LLM’s responses to ensure they remain within desired ethical and practical parameters.

\begin{figure}[h]
\centering

\begin{tcolorbox}[colback=white!10!white, colframe=blue!75!black, title=Instruction]
Plan a 10-day cultural tour of Japan, including visits to Tokyo, Kyoto, and Nara during cherry blossom season.
\end{tcolorbox}

\begin{tcolorbox}[colback=white!10!white, colframe=green!55!black, title=Alignment in Travel Planning]
Alignment involves ensuring that the LLM's outputs adhere to user preferences (e.g., budget, safety), ethical standards, and contextual appropriateness. Effective alignment strategies include technical measures such as adjusting model parameters and training data, and policy measures such as incorporating feedback loops that allow users to report and rectify misaligned behavior. A knowledge graph can further support alignment by embedding a structured understanding of ethical norms and user preferences directly into the model’s reasoning processes.
\begin{itemize}
    \item \textbf{Accommodation Preferences:} Aligns hotel selections to provide views of cherry blossoms while also being located near cultural sites, catering to the user's preferences for scenic and enriching experiences.
    \item \textbf{Safety and Accessibility:} Ensures that all recommended activities and accommodations meet general safety standards and are accessible, considering specific user needs like accessibility features for disabled travelers.
    \item \textbf{Personalized Recommendations:} Offers customized recommendations based on the user's dietary preferences and budget constraints, enhancing the user's overall experience and satisfaction.
\end{itemize}
\end{tcolorbox}
\caption{Illustration of Alignment in LLM-based Travel Planning}
\label{fig:alignment_travel_planning}
\end{figure}
\section{Conclusion}

We introduced a neurosymbolic framework to enhance the instructability of generative AI, with LLMs as a prime example, addressing the limitations of traditional instruction tuning approaches. By integrating symbolic task planners with neural semantic parsers and neurosymbolic executors, we discuss how to achieve superior task decomposition, semantic grounding, and execution reliability. Knowledge graphs further enrich this integration, providing essential real-world context and ensuring alignment with ethical standards and user expectations.

\section*{Acknowledgements}
This research was supported in part by NSF Awards \#2335967 ``EAGER: Knowledge-guided neurosymbolic AI with guardrails for safe virtual health assistants'' and \#2119654, ``RII Track 2 FEC: Enabling Factory to Factory (F2F) Networking for Future
Manufacturing''. Opinions are those of authors and not the sponsor.

\section*{Authors}

\noindent \textbf{Amit Sheth} is the NCR Chair, and a professor; he founded the university-wide AI Institute of South Carolina (AIISC) in 2019. He received the 2023 IEEE-CS Wallace McDowell award and is a fellow of IEEE, AAAI, AAIA, AAAS, and ACM. Contact him at: \texttt{amit@sc.edu} 

\noindent \textbf{Vishal Pallagani} is a Ph.D. student at AIISC. His research focuses on the intersection of automated planning, generative AI, and their integration in building neurosymbolic solvers. Contact him at: \texttt{vishalp@email.sc.edu}

\noindent \textbf{Kaushik Roy}
is a Ph.D. candidate at the AIISC. His research focuses on developing neurosymbolic methods for declarative and process knowledge-infused learning, reasoning, and sequential decision-making, particularly emphasizing social good applications. He has a strong track record of publications related to this topic. Contact him at: \texttt{kaushikr@email.sc.edu} 

\bibliographystyle{IEEEtran}
\bibliography{references}

\end{document}